\documentclass[10pt,twocolumn,letterpaper]{article}

\usepackage{cvpr}    

\usepackage[dvipsnames,table]{xcolor}
\usepackage{anyfontsize}
\usepackage{array}
\usepackage{booktabs}
\usepackage[export]{adjustbox}

\definecolor{cvprblue}{rgb}{0.21,0.49,0.74}
\usepackage[pagebackref,breaklinks,colorlinks,citecolor=cvprblue]{hyperref}
\usepackage{multirow}
\usepackage{subcaption}

\title{Robust Confidence Intervals in Stereo Matching using Possibility Theory}

\author{Roman Malinowski\\
CNES, CS, UTC\\
18 avenue E. Belin, Toulouse, France\\
{\tt\small roman.malinowski@utc.fr}
\and
Emmanuelle Sarrazin\\
Centre National d'Etudes Spatiales (CNES)\\
18 avenue E. Belin, Toulouse, France\\
{\tt\small emmanuelle.sarrazin@cnes.fr}
\and
Loïc Dumas\\
CS\\
5 rue Brindejonc des Moulinais, Toulouse, France\\
{\tt\small loic.dumas@csgroup.eu}
\and
Emmanuel Dubois\\
Centre National d'Etudes Spatiales (CNES)\\
18 avenue E. Belin, Toulouse, France\\
{\tt\small emmanuel.dubois@cnes.fr}
\and
Sébastien Destercke\\
UTC\\
Université de Technologie de Compiègne (UTC)\\
Avenue de Landshut, Compiègne, France\\
{\tt\small sebastien.destercke@utc.fr}
}

\begin{document}
\maketitle

\begin{abstract}
We propose a method for estimating disparity confidence intervals in stereo matching problems. Confidence intervals provide complementary information to usual confidence measures. To the best of our knowledge, this is the first method creating disparity confidence intervals based on the cost volume. This method relies on possibility distributions to interpret the epistemic uncertainty of the cost volume. Our method has the benefit of having a white-box nature, differing in this respect from current state-of-the-art deep neural networks approaches. The accuracy and size of confidence intervals are validated using the Middlebury stereo datasets as well as a dataset of satellite images. This contribution is freely available on GitHub.
\end{abstract}

\begin{figure}[hb!]
\centering\begin{subfigure}{0.5\linewidth}
  \includegraphics[width=1\linewidth]{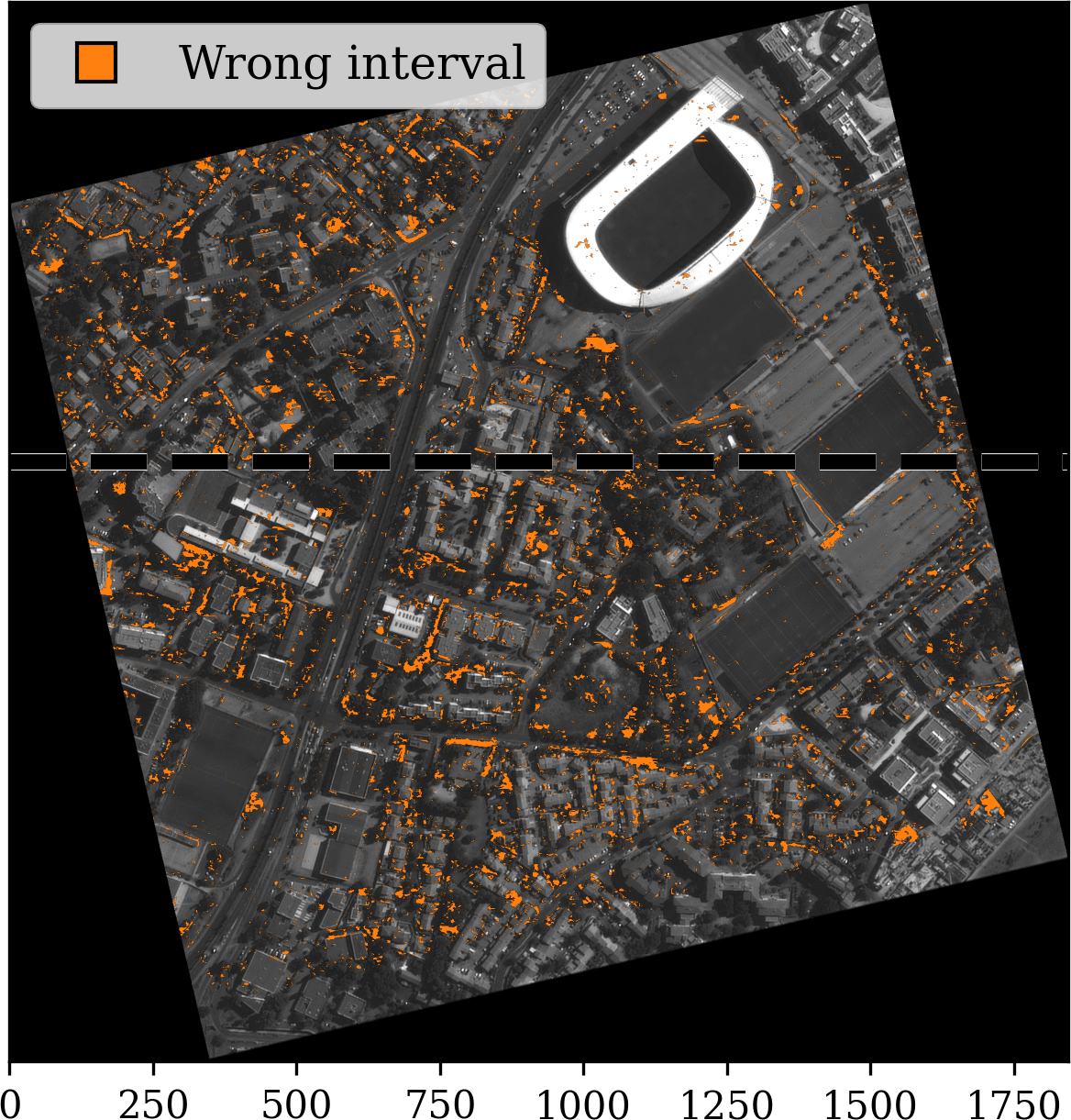}
  \caption{Left epipolar image}
    \label{fig:fig1a}
\end{subfigure}
\begin{subfigure}{1\linewidth}
    \centering
  \includegraphics[width=1\linewidth]{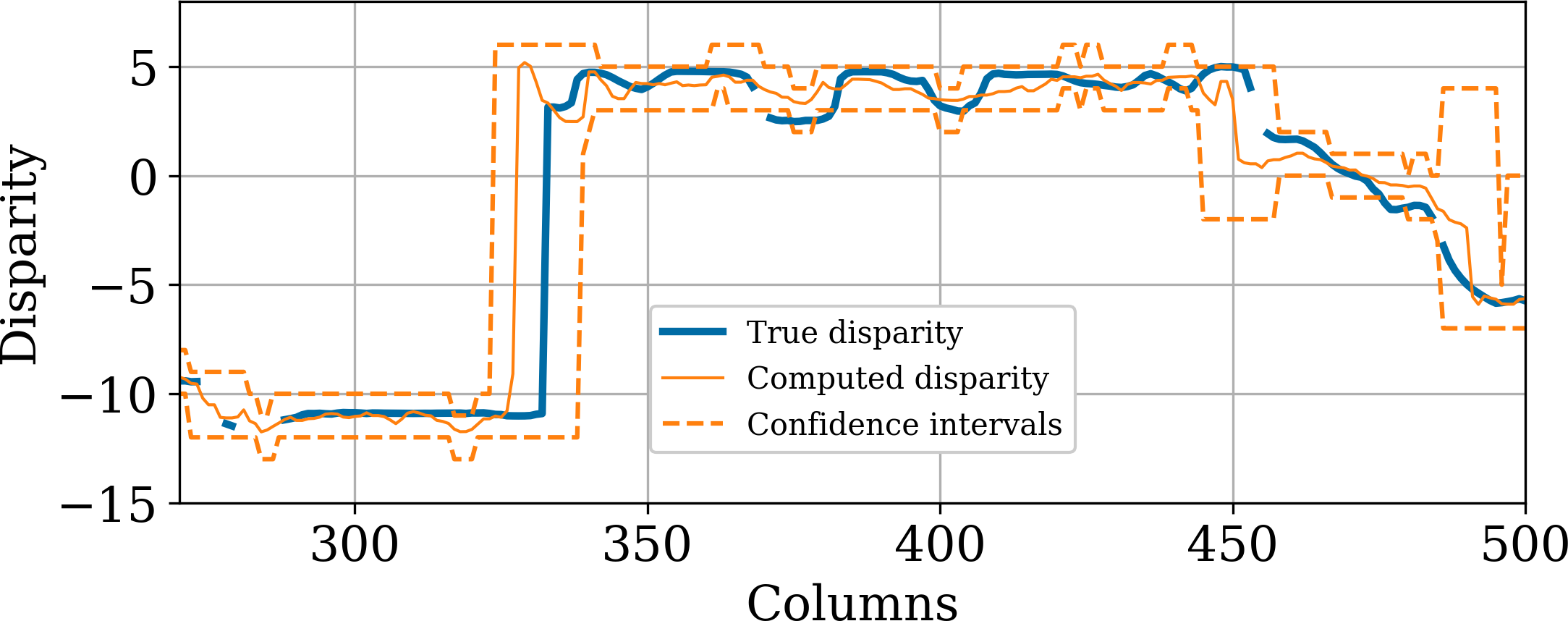}
    \caption{Confidence Intervals}
  \label{fig:fig1b}
\end{subfigure}
\caption{Example of intervals on a image of the city of Montpellier, France. \cref{fig:fig1a} presents the left image, colored pixels indicate wrong interval locations. \cref{fig:fig1b} contains confidence intervals along a section of the dashed line in \cref{fig:fig1a}}
   \label{fig:fig1}
\end{figure}

\section{Introduction}
Stereo matching is used as a mean to estimate the depth of a scene in numerous applications, ranging from autonomous driving to Earth observation \cite{geiger_are_2012,marti_mapping_2016}. With the growing availability of satellite imagery \cite{melet_co3d_2020}, many stereo algorithms have been proposed to perform 3D reconstruction from remote sensing images \cite{shean_automated_2016, franchis_automatic_2014, rupnik_micmac_2017, youssefi_cars_2020}. All those algorithms contain a dense matching step, which consists in determining the displacement of every pixel, called disparity, between the different images. Such algorithms usually start by computing the similarity or sets of features between pixels in the form of a cost volume, from which the disparity can then be deduced \cite{scharstein_taxonomy_2001, laga_survey_2022}. Point clouds are retrieved from the disparity, which can themselves be converted into a mesh or a digital surface model. 

Estimating the confidence in the disparity estimation is crucial in many applications, and can lead to the improvement of overall results \cite{hollmann_geometry-based_2020, sarrazin_ambiguity_2021}. In the context of 3D reconstruction from satellite imagery, it can even be propagated to be provided as the confidence related to the final 3D product. As such, it has become an important research topic \cite{hu_quantitative_2012, poggi_confidence_2021}. There are two aspects to the uncertainty regarding disparity computation: how confident we are in the disparity prediction, and what would be the magnitude of the potential error. Although state-of-the-art confidence measures reliably indicate how likely a predicted disparity is to be correct, they do not indicate the extent of the potential error. Those two notions are linked, but are not the same: it could be that a prediction is made with high confidence but would have a large associated error if wrong (meaning there would be a great gap between the predicted and the true disparity in case of an error). Similarly, a prediction could be made with low confidence, but the set of possible disparities is restricted to few values close to the predicted disparity. Estimating the magnitude of the error and providing sets of possible disparities bring additional information that could help users to improve the disparity map, similarly to current work with classical confidence measures \cite{spyropoulos_learning_2014, poggi_learning_2016}. 

In this article, we present a method for creating robust disparity confidence intervals on the disparity estimation. The intervals will be propagated in future applications to produce confidence intervals on digital surface models, but this lies outside the scope of this paper and thus will not be covered here. We design our method so that it can be fully integrated in classical 3D pipelines \cite{shean_automated_2016, franchis_automatic_2014, rupnik_micmac_2017, youssefi_cars_2020} using a cost-volume based stereo matching algorithm depicted in Scharstein \etal \cite{scharstein_taxonomy_2001}. To the best of our knowledge, this is the first approach providing disparity confidence intervals for stereo matching problems. For each pixel of the reference image, we give a lower and upper displacement of its position in the target image as in \cref{fig:fig1}. We aim for an accuracy of $90\%$, using robust uncertainty models called \textit{possibility distributions}. We think this additional information about the magnitude of the error gives a deeper understanding of the uncertainty in stereo matching algorithms. No training is required to produce confidence intervals, also sparing the method from classical criticisms regarding black-box aspects, in the sense that all processing can be followed and monitored. Additionally, we detail precautions that must be taken when applying some post-processing steps \cite{scharstein_taxonomy_2001} to the disparity map so that it stays consistent with the confidence intervals. Our method for creating confidence intervals can be summarized as follows:
\begin{enumerate}
	\item Computation of the matching cost volume and confidence measure
	\item Transformation of matching cost curves into possibility distributions
	\item Deduction of disparity intervals by taking $\alpha$-cuts on the possibility distributions
 	\item Filtering of intervals while maintaining consistency with the disparity map
	\item Statistical regularization of intervals in low-confidence zones
\end{enumerate}

\Cref{sec:related_works} contains an overview of current work regarding stereo algorithms, confidence measures, and uncertainty models used in this paper. \Cref{sec:intervals} details the method for constructing confidence intervals. Finally, the confidence intervals robustness and size are validated in \Cref{sec:evaluation} using images from the Middlebury dataset and from a dataset of high-resolution optical satellite images of various landscapes around the city of Montpellier. The code is freely available on GitHub: \url{https://github.com/CNES/Pandora}.
\section{Related Works}\label{sec:related_works}
\subsection{Stereo matching}
Our method is designed for classical 3D reconstruction pipelines from remote sensing imagery \cite{shean_automated_2016, franchis_automatic_2014, rupnik_micmac_2017, youssefi_cars_2020}. Those pipelines retrieve stereo information by means of dense matching algorithms, mainly falling into two main categories: classical approaches following the steps established by Scharstein \etal \cite{scharstein_taxonomy_2001}, and full deep-learning approaches. Classical approaches usually contain the following steps: matching cost computation, cost aggregation, disparity computation and disparity refinement. On the other hand, deep-learning approaches provide strong results (see \cite{laga_survey_2022} for details), but are prone to generalization issues when using images differing from the training dataset, especially in the context of satellite imagery  \cite{mari_disparity_2022}. Obtaining ground truth data of various landscape can prove difficult, we therefore focus here on classical approaches as we aim to produce a general and robust method for creating confidence intervals. We consider two similarity functions, the Census cost function \cite{goos_non-parametric_1994}, and the MC-CNN cost function \cite{zbontar_stereo_2016} learned using convolutional neural networks. Those similarity functions are \textit{minimitive}, meaning that a low value indicates a strong similarity of the compared patches. Both handcrafted and learned similarity functions are considered here to highlight the generic nature of our method for creating intervals. The cost volume obtained using those functions is regularized using the semi-global matching (SGM) methods \cite{hirschmuller_accurate_2005, facciolo_mgm_2015}, used in other state-of-the-art methods \cite{chebbi_deepsim-nets_2023}.

\begin{figure}[t]
  \centering
  \includegraphics[width=1\linewidth]{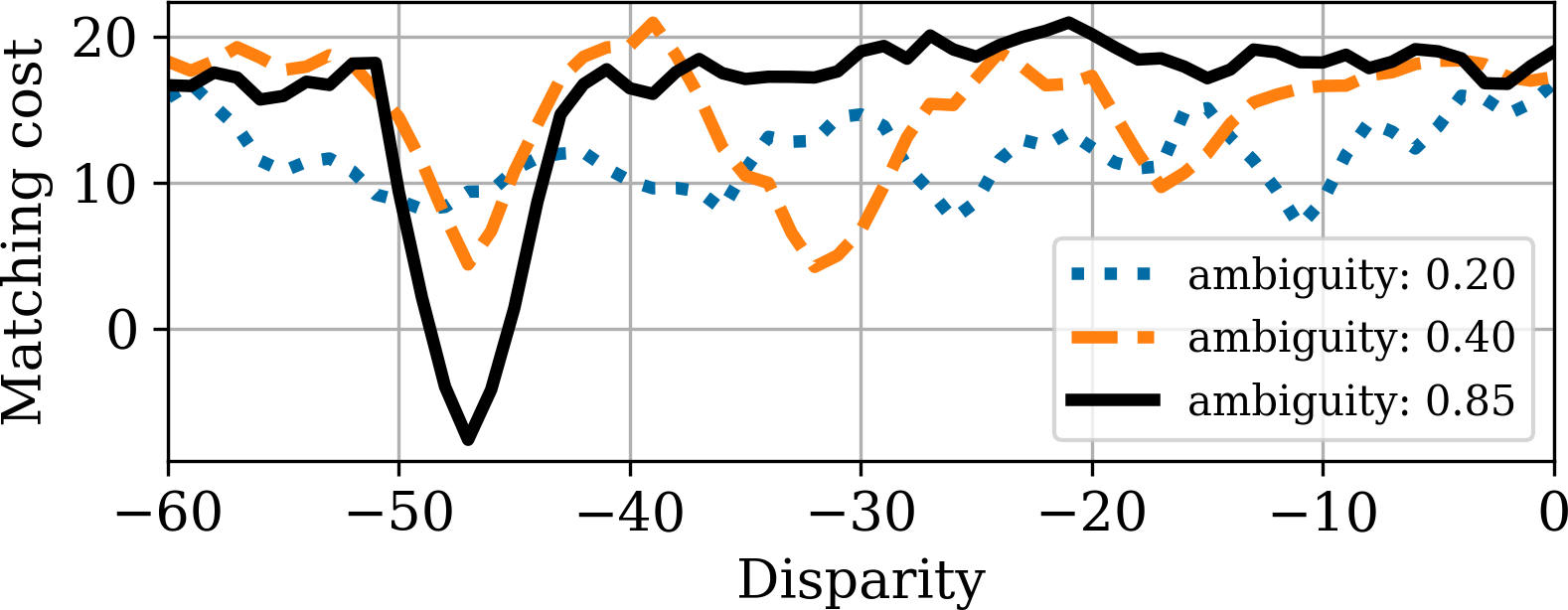}
  \caption{Example of three MC-CNN cost curves with different confidences from ambiguity.}
   \label{fig:ambiguity}
\end{figure}

Numerous confidence measures have been proposed regarding the disparity \cite{hu_quantitative_2012}, handcrafted using the properties from the reference image, the cost volume or the disparity map itself. Learning-based methods constitute the majority of state of the art confidence measures \cite{poggi_confidence_2021}. We can mention for instance deep learning methods estimating confidence using CNNs on the cost volume \cite{mehltretter_cnn-based_2019} or on the disparity map \cite{poggi_learning_2016}, and random forests on handcrafted confidence measures \cite{gouveia_confidence_2015}. We refer to \cite{poggi_confidence_2021, hu_quantitative_2012} for more in-depth details. Estimating the confidence supporting the disparity map can lead to new strategies to improve the overall results \cite{dumas_improving_2022, tosi_leveraging_2019, hollmann_geometry-based_2020}. In this work, we use use a confidence measure computed from the matching cost volume, called \textit{ambiguity} \cite{sarrazin_ambiguity_2021}.

In \cite{chen_learning_2023}, authors estimate the magnitude of the absolute symmetric error using a MLP. Their work demonstrate the interest in estimating the magnitude of the error as a complement to confident measures, and we push this idea even further. Indeed, estimating the absolute error provides valuable information on the magnitude of the error, but does not indicate \textit{where} the correct disparity should be, i.e. if it is probably higher or lower than the predicted one.

We therefore propose to answer this question using confidence intervals on the disparity. Our method also differs as it can be plugged on a vast range of cost-volume based stereo algorithms with \textit{winner-takes-all} strategy. Additionally, our method does not rely on the accuracy of the algorithm used for the disparity prediction. In contrast, the network in \cite{chen_learning_2023} 
specifically takes as input $4$ disparities at various resolutions estimated by a 3D CNN \cite{guo_group-wise_2019}, limiting its applicability. To the best of our knowledge, this is the only work that estimates the disparity error in a manner similar to ours. The restricted setting in which they work also means that we cannot compare our two approaches. 

\subsection{Possibility distributions}
To create confidence intervals, we consider using \textit{possibility distributions}, closely related \textit{fuzzy sets} \cite{dubois_random_1991}, as uncertainty models. This allows to correctly represent \textit{epistemic} uncertainty, i.e., due to the partial nature of available information \cite{baudrit_joint_2007}. In stereo matching problems, errors are mostly due to epistemic uncertainty. Indeed, similarity functions evaluate how much two patches are alike based on given or learned features, and there exists some uncertainty regarding how well this similarity indicates a match between corresponding pixels. Using possibility distributions aims to address the downsides of probability distributions regarding epistemic uncertainty \cite{walley_statistical_1991}. Possibility distributions are well-suited to model an expert's opinion on the uncertainty of an imprecise observation, for instance in the context of groundwater contamination \cite{bardossy_l-_1995, baudrit_joint_2007}. In the context of stereo matching, a regularized cost curve using SGM can be seen as an expert evaluating if two patches are homologous or not, based on their features and global properties of the cost volume. The use of possibility distributions allows to benefit from the advanced state of knowledge of IP towards robust estimation of uncertainty. 

Formally, a possibility distribution is defined as a mapping $\pi:\Omega\rightarrow[0,1]$ verifying:
\begin{equation}
	\exists \omega\in\Omega, \pi(\omega)=1 \label{eq:possibility}
\end{equation}
where $\Omega$ is the set of possible observed states. $\pi$ represents the degree of possibility of an event $\omega$, $\pi(\omega)=1$ meaning that $\omega$ is fully possible, and $\pi(\omega)=0$ meaning that $\omega$ is impossible. Possibility distributions can be used to define the envelope of a convex set of probability distributions $\mathbb{P}$ \cite{dubois_when_1992}:
\begin{equation}
    \mathbb{P}=\{P:2^\Omega\rightarrow[0,1]~|~P(E)\leqslant\sup_{\omega\in E}\pi(\omega)\} \label{eq:credal_set}
\end{equation}
where $P$ is a probability distribution on the power set $2^\Omega$ of $\Omega$.

Alongside possibility distributions are often defined $\alpha$-cuts $\mathcal{C}^{\pi}_\alpha$, which will be used for constructing the confidence intervals:
\begin{equation}
	C_\alpha^\pi=\{\omega\in\Omega~|~\pi(\omega)\geqslant\alpha\}\label{eq:alpha_cut}
\end{equation}
$\alpha$-cuts are the maximal sets whose possibility is at least equal to $\alpha$ for every $\omega\in\Omega$. From a probabilistic point of view, $\alpha$-cuts are composed of all $\omega\in\Omega$ for which there is a $P\in\mathbb{P}$ from \cref{eq:credal_set} such that $P(\omega)\geqslant\alpha$:
\begin{equation}
    C_\alpha^\pi=\{\omega~|~\exists P\in\mathbb{P},~P(\omega)\geqslant \alpha\}\label{eq:alpha_proba}
\end{equation}
\section{Creation of Disparity Confidence Intervals}\label{sec:intervals}
In the following, we consider that the images have been resampled in epipolar geometry so that the displacement of a pixel between left $I_L$ and right $I_R$ images can only occur horizontally in a given disparity range $\mathcal{D}=[d_{min}, d_{max}]$. In this setting, a pixel of the left image $p=(i,j)$ with a disparity $d\in\mathcal{D}$ is matched to the pixel $q=(i, j+d)$ of the right image with a cost $C_V(i,j,d)$.

\subsection{From Cost Curves to Possibility Distributions}
We use possibility distributions to model the epistemic uncertainty associated with similarity functions, in the same way that an expert would state an opinion for every pixel on which disparities are more likely to be the correct ones.

In order to construct a possibility distribution consistent with a cost curve, it is first necessary to normalize the cost curve to ensure its values lies in $[0,1]$. Normalization can be done using the minimum and maximum values of the matching cost volume after SGM regularization. Given a pixel $p=(i,j)$, its cost curve is normalized as follows:
\begin{eqnarray}
	f_{i,j}^{norm}(d)=\frac{C_V(i,j,d)-\max C_V}{\min C_V-\max C_V}
\end{eqnarray}
The $\min$ and $\max$ operators of standard normalization are reversed as the possibility needs to be maximal when the cost function is minimal. $f_{i,j}^{norm}$ is not yet a possibility distribution as it does not necessarily verify \cref{eq:possibility}. As noticed in \cref{eq:possibility}, we can interpret it as a form of inconsistency, in the sense that there is no disparity value such that the two patches of pixels perfectly match. Assuming $p=(i,j)$ has an homologous pixel in the right image, we can restore consistency through normalisation~\cite{oussalah_normalization_2002} 
, resulting in a possibility distribution $\pi_{i,j}(d):\mathcal{D}\rightarrow[0,1]$:
\begin{equation}
	\pi_{i,j}(d) = f_{i,j}^{norm}(d) + 1 - \max_{d\in\mathcal{D}}f_{i,j}^{norm}(d) \label{eq:constant_norm}
\end{equation}

\begin{figure}[t]
  \centering
  \includegraphics[width=1\linewidth]{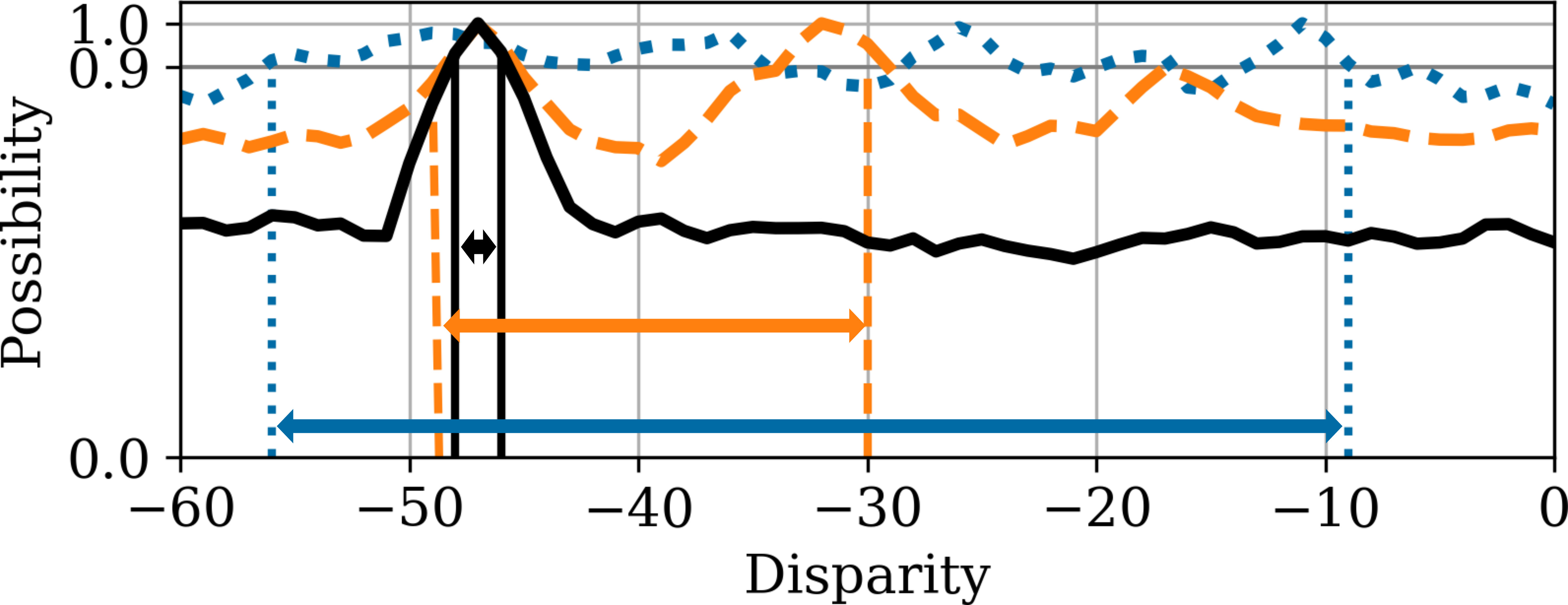}
  \caption{The possibility distributions obtained from the cost curves of \cref{fig:ambiguity}. The arrows and vertical lines indicate the disparity intervals obtained with $\alpha=0.9$.}
   \label{fig:possibility}
\end{figure}

Examples of cost curved transformed into possibility distributions are presented in \cref{fig:ambiguity,fig:possibility}. Another way of verifying \cref{eq:possibility} would have been to obtain $f_{i,j}^{norm}$ using the $\min$ and $\max$ operators on the cost curve instead of the whole cost volume. Doing so would have artificially accentuated the differences between the matching costs \cite{oussalah_normalization_2002}. Using \cref{eq:constant_norm} instead keeps the curvature of the cost curve.

\subsection{Deducing Intervals from Alpha-cuts}
Having defined possibility distributions, we now look to define a set of possible disparities verifying our $90\%$ confidence objective. Every possibility distribution defines a set of probability distributions $\mathbb{P}$ using \cref{eq:credal_set}. Disparities $d$ for which every probability measure in $\mathbb{P}$ evaluated on $d$ are lower than $0.9$ are deemed to be unlikely. According to \cref{eq:alpha_proba}, considering disparities $d$ for which there is a probability $P\in\mathbb{P}$ such that $P(d)\geqslant 0.9$ is equivalent to consider the $\alpha$-cut $C^{\pi_{i,j}}_\alpha$ with $\alpha=0.9$. Different values of $\alpha$ reflect different levels of confidence. We compared different values of $\alpha$ in the ablation study presented in \cref{tab:baseline_alpha}.

In general, $\alpha$-cuts are sets and not intervals. We are able to define a single confidence interval $I_\alpha(i,j)$ from an $\alpha$-cut by taking its extrema:
\begin{equation}
	I_\alpha(i,j)=[\min C^{\pi_{i,j}}_\alpha, \max C^{\pi_{i,j}}_\alpha] \label{eq:confidence_interval}
\end{equation}
Switching from $C^{\pi_{i,j}}_\alpha$ to the interval $I_\alpha(i,j)$ reduces the amount of information available by adding disparities with low possibilities to our considered set of disparities. However, as only two interval bounds need to be considered instead of every disparity of $C^{\pi_{i,j}}_\alpha$, our solution consumes less memory, facilitates further processing and is easier to understand for users. The level of confidence of $I_\alpha(i,j)$ is also guaranteed to be at least equal to that of $C^{\pi_{i,j}}_\alpha$ as $C^{\pi_{i,j}}_\alpha\subseteq I_\alpha(i,j)$. Examples of confidence intervals on disparities are presented in \cref{fig:possibility}. 

\subsection{Refinement and Filtering with Intervals}
In most stereo pipelines, the output disparity map deduced from the matching cost volume is being post-processed. Namely, sub-pixel refinement and filtering steps are usually applied to improve overall results. The confidence intervals need to be processed accordingly to maintain their coherence with the disparity map. 

Sub-pixel refinement is taking into account by slightly extending the confidence intervals in the case where the predicted disparity $d_{i,j}$ is one of the interval bounds. For clarity, we refer to the lower and upper bounds of an interval $I$ as $\underline{I}$ and $\overline{I}$ respectively. Confidence intervals are modified as follows:
\begin{eqnarray}
    \text{if }d_{i,j}=\min C^{\pi_{i,j}}_\alpha, \underline{I}_\alpha(i,j) = \min C^{\pi_{i,j}}_\alpha - 1\\
    \text{if }d_{i,j}=\max C^{\pi_{i,j}}_\alpha, \overline{I}_\alpha(i,j) = \max C^{\pi_{i,j}}_\alpha + 1
\end{eqnarray}
This interval extension is coherent with different methods of interpolation, like parabolic-fit for instance. V-fit refinement \cite{haller_real-time_2010} is used in our experiments.

Similarly, multiple filtering of the disparity map can be considered. We use a median filter in our experiments, which is a popular method in many stereo algorithms \cite{scharstein_taxonomy_2001, facciolo_mgm_2015}. Using a median filter modifies the disparity map and might create inconsistencies with the confidence intervals. However, it is possible to demonstrate that for every set of pixels $\{p(i,j)\}$ and confidence intervals $\{I_\alpha(i,j)\}$ verifying $\forall (i,j), p(i,j)\in I_\alpha(i,j)$ then:

\begin{eqnarray}
    median(\{\underline{I}_\alpha(i,j)\}) \leqslant median(\{p(i,j)\})\label{eq:median_1}\\
    median(\{p(i,j)\}) \leqslant median(\{\overline{I}_\alpha(i,j)\})\label{eq:median_2}
\end{eqnarray}
Previous equations mean that the median filter can be applied independently to the disparity map and the confidence interval bounds while still maintaining their coherence.

\subsection{Regularization in Low Confidence Areas}\label{subsec:regularization}

Despite running post-processing steps, confidence interval performances heavily depend on the quality of the similarity function used. Near surface discontinuities, SGM algorithm sometimes struggle to correctly detect disparity changes due to the continuity constraint. A shift between the predicted and the true disparity can be observed, which induces biases in the cost curve and challenges the interpretation of the cost curve as an expert's opinion. To overcome this limitation, confidence intervals are processed with a more pessimistic approach in those areas.

Low confidence areas are detected using confidence measures. As such, our approach is complementary to classical confidence estimation approaches. We use the confidence from ambiguity measure \cite{sarrazin_ambiguity_2021} as it presents the advantage of being both explainable and efficient for this task. This confidence measure aims to represent the difficulty to single out a disparity value as the correct disparity. The higher the confidence from ambiguity, the easier it is easy to identify the correct disparity, whereas a value near $0$ indicates that numerous patches present the same locally minimal similarity. Examples of cost curves with different values of ambiguities can be found in \cref{fig:ambiguity}. The ambiguity measure is computed pixel-wise, and thus can present strong spatial variations in low confidence zones. To compensate this effect, we first smooth the ambiguity map using a $1\times5$ $\min$ convolution kernel, and pixels whose ambiguity is under a threshold $\tau$ are considered to be in low confidence zones:
\begin{eqnarray}
    amb_{smooth}(i,j) = \min_{-2\leqslant k\leqslant 2} amb(i, j+k)\\
    \text{Low confidence if }~amb_{smooth}(i,j)\leqslant\tau \label{eq:low_confidence}
\end{eqnarray}
We fix empirically $\tau=0.6$ for our experiments with ambiguity, as it does not seem to depend on the similarity function used.

Low confidence areas usually correspond to regions which possess a strong disparity variation. The intervals in low confidence areas might be extended to the minimal and maximal bounds found in the area, but experiments show this approach is too pessimistic. Indeed, a single unnecessary large confidence interval would penalize the whole area. Instead, we advocate for a statistical approach by extending the intervals using quantiles instead of the the minimal and maximal bounds. This approach has the advantage of being more robust to outliers.

\begin{figure}[t]
  \centering
  \includegraphics[width=\linewidth]{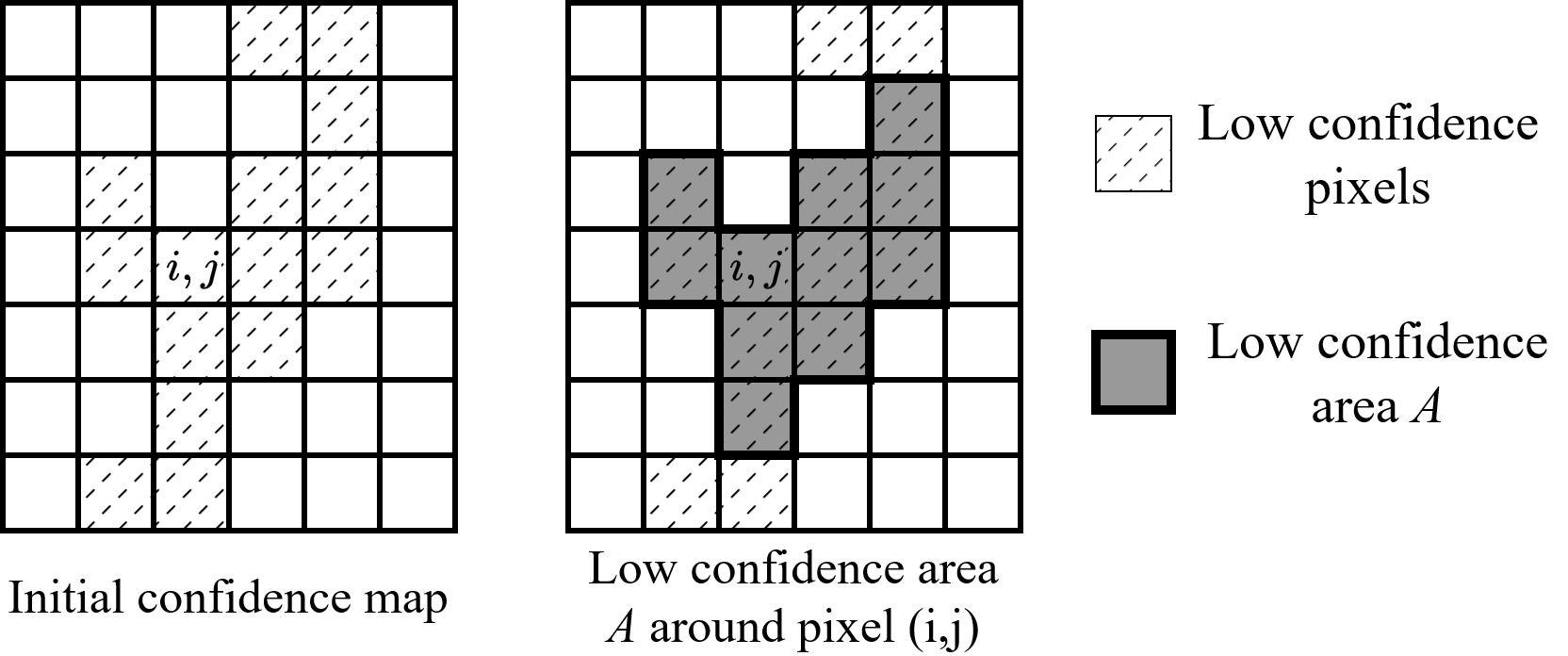}
  \caption{Low confidence areas with $l=2$.}
   \label{fig:low_confidence_area}
\end{figure}

\begin{figure}[ht]
\centering\begin{subfigure}{0.5\linewidth}
  \includegraphics[width=1\linewidth, left]{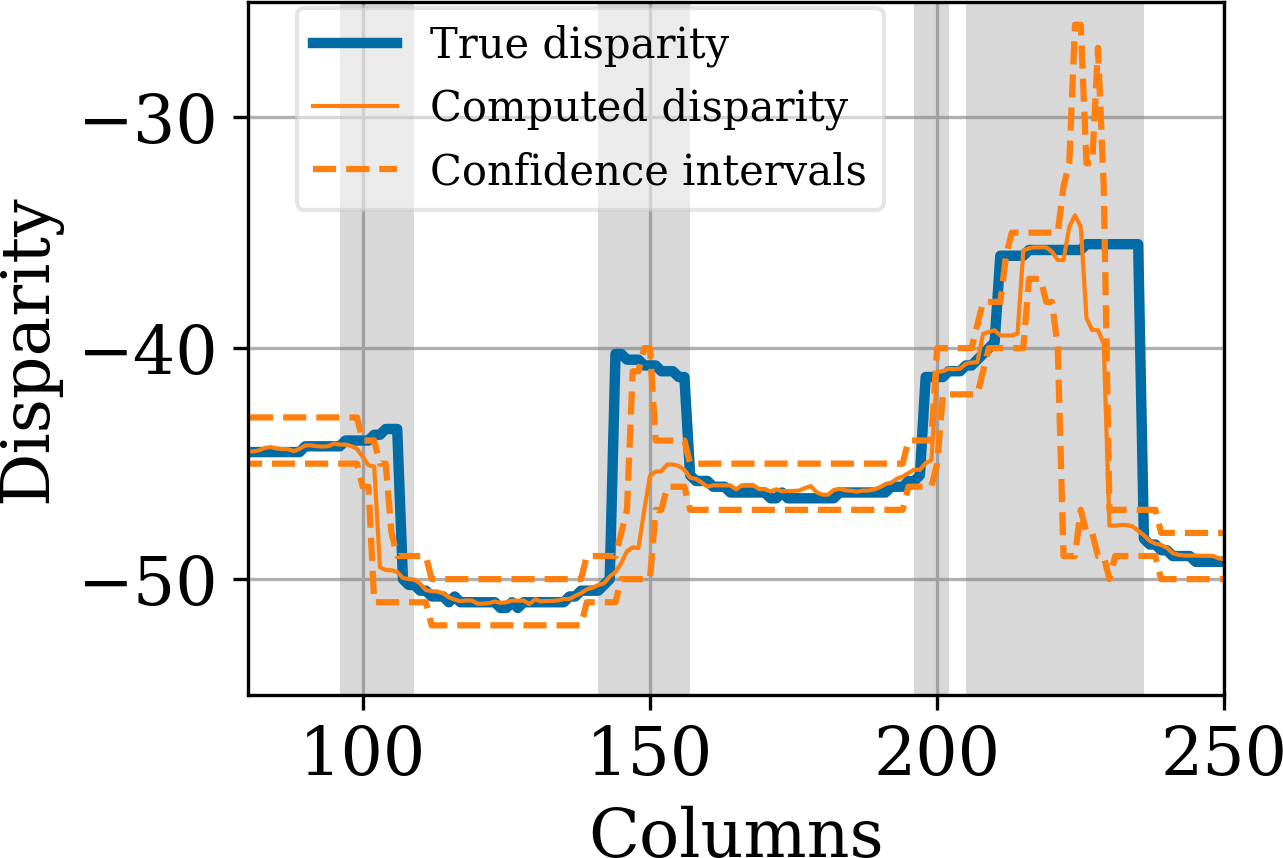}
  \caption{Intervals without regularization}
  \label{fig:no_reg}
\end{subfigure}\begin{subfigure}{0.5\linewidth}
    \centering
  \includegraphics[width=1\linewidth, right]{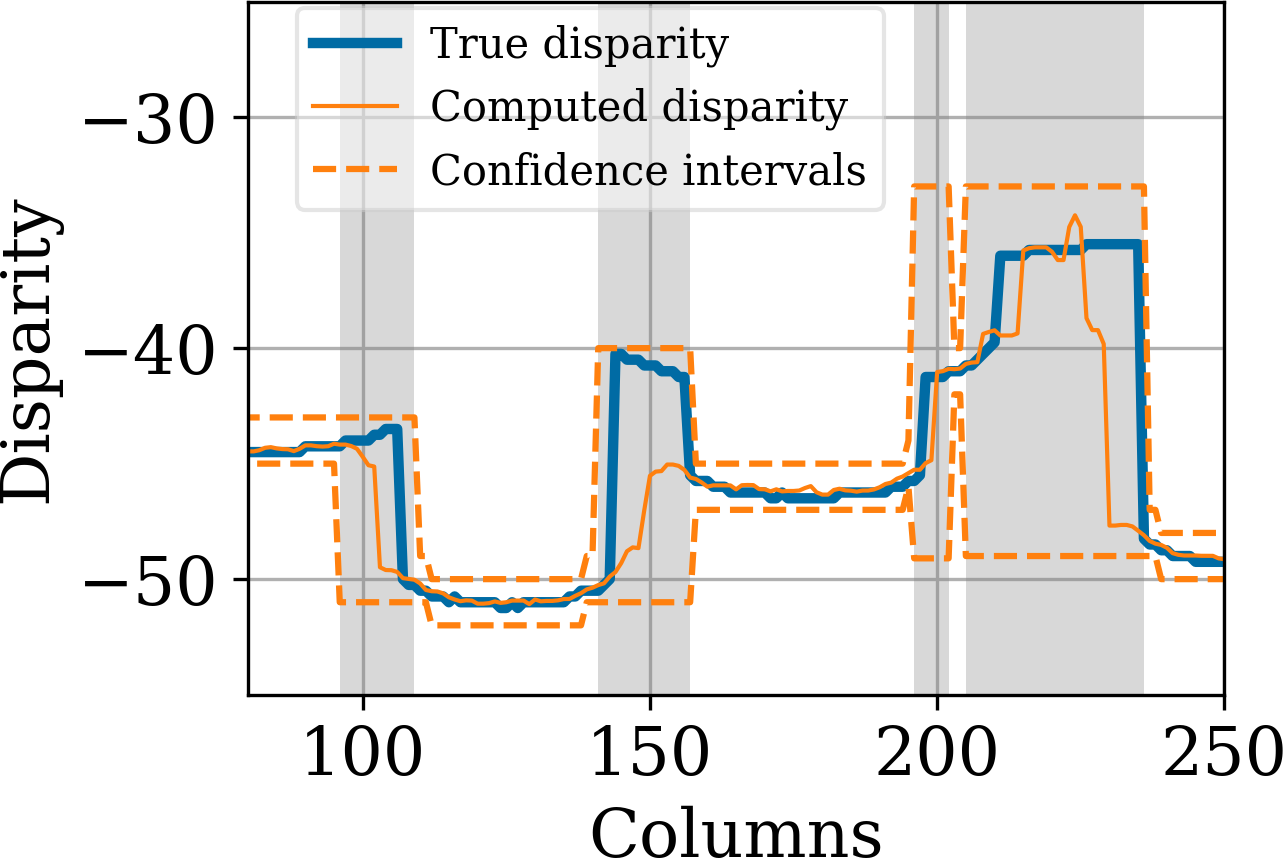}
    \caption{Interval with regularization}
  \label{fig:reg}
\end{subfigure}
\caption{Intervals without (\cref{fig:no_reg}) and with (\cref{fig:reg}) regularization from Middlebury's $cones$. CENSUS cost function is used. Areas with low confidence are indicated in gray.}
   \label{fig:regularization_example}
\end{figure}

For each low confidence pixel, we determine the maximal set $A$ of contiguous low confidence pixels within $l$ lines above and bellow, as presented for $l=2$ in \cref{fig:low_confidence_area}. Experiments show that $l=2$ allows the set $A$ to contain enough pixels to be statistically relevant, while maintaining a relatively low computation time. Using values higher than $2$ does not improve the results. Once $A$ has been established, the lower interval bound of the low confidence pixel is set to the $10^{th}$ quantile of the lower interval bounds in $A$. The same procedure is applied to the upper bounds with the $90^{th}$ quantile. Examples of intervals with and without regularization along a row of a scene are presented in \cref{fig:regularization_example}. When no regularization is applied, errors occur for intervals in low confidence areas. 

\section{Evaluation}\label{sec:evaluation}

\newcolumntype{?}[1]{!{\vrule width #1}}
\setlength{\tabcolsep}{3pt} 
\renewcommand{\arraystretch}{1.4} 
\begin{table*}
\centering{
\begin{tabular}{ |c?{1.5pt}c|c?{1.2pt}c|c?{1.5pt}c|c?{1.2pt}c|c?{1.5pt}c|c?{1.2pt}c|c?{1.2pt}c|c?{1.5pt}}
\cline{2-15}
\multicolumn{1}{c?{1.5pt}}{} & \multicolumn{4}{c?{1.5pt}}{Global} & \multicolumn{4}{c?{1.5pt}}{High confidence areas} & \multicolumn{6}{c?{1.5pt}}{Low confidence areas}\\
\cline{2-15}
\multicolumn{1}{c?{1.5pt}}{} & \multicolumn{2}{c?{1.2pt}}{\fontsize{8}{9.6}\selectfont Accuracy $\uparrow$} & \multicolumn{2}{c?{1.5pt}}{\fontsize{8}{9.6}\selectfont Relative Size $\downarrow$} & \multicolumn{2}{c?{1.2pt}}{\fontsize{8}{9.6}\selectfont Accuracy $\uparrow$} & \multicolumn{2}{c?{1.5pt}}{\fontsize{8}{9.6}\selectfont Relative Size $\downarrow$} & \multicolumn{2}{c?{1.2pt}}{\fontsize{8}{9.6}\selectfont Accuracy $\uparrow$} & \multicolumn{2}{c?{1.2pt}}{\fontsize{8}{9.6}\selectfont Relative Size $\downarrow$} & \multicolumn{2}{c?{1.5pt}}{\fontsize{8}{9.6}\selectfont Overestimation $\downarrow$}\\
\hline
Dataset & {\fontsize{7}{8.4}\selectfont CENSUS} & {\fontsize{7}{8.4}\selectfont MCCNN} & {\fontsize{7}{8.4}\selectfont CENSUS} & {\fontsize{7}{8.4}\selectfont MCCNN} & {\fontsize{7}{8.4}\selectfont CENSUS} & {\fontsize{7}{8.4}\selectfont MCCNN} & {\fontsize{7}{8.4}\selectfont CENSUS} & {\fontsize{7}{8.4}\selectfont MCCNN} & {\fontsize{7}{8.4}\selectfont CENSUS} & {\fontsize{7}{8.4}\selectfont MCCNN} & {\fontsize{7}{8.4}\selectfont CENSUS} & {\fontsize{7}{8.4}\selectfont MCCNN} & {\fontsize{7}{8.4}\selectfont CENSUS} & {\fontsize{7}{8.4}\selectfont MCCNN}\\
\hline\hline
\cellcolor{lightgray!10}$2003$ & \cellcolor{lightgray!10}$\textbf{0.973}$ & \cellcolor{lightgray!10}$0.954$ & \cellcolor{lightgray!10}$\textbf{0.033}$ & \cellcolor{lightgray!10}$\textbf{0.033}$ & \cellcolor{lightgray!10}$\textbf{0.983}$ & \cellcolor{lightgray!10}$0.968$ & \cellcolor{lightgray!10}$\textbf{0.033}$ & \cellcolor{lightgray!10}$\textbf{0.033}$ & \cellcolor{lightgray!10}$\textbf{0.942}$ & \cellcolor{lightgray!10}$0.89$ & \cellcolor{lightgray!10}$\textbf{0.183}$ & \cellcolor{lightgray!10}$0.233$ & \cellcolor{lightgray!10}$\textbf{0.165}$ & \cellcolor{lightgray!10}$0.182$ \\ 
\cellcolor{gray!30}$2005$ & \cellcolor{gray!30}$0.963$ & \cellcolor{gray!30}$\textbf{0.971}$ & \cellcolor{gray!30}$\textbf{0.026}$ & \cellcolor{gray!30}$0.038$ & \cellcolor{gray!30}$0.969$ & \cellcolor{gray!30}$\textbf{0.973}$ & \cellcolor{gray!30}$\textbf{0.026}$ & \cellcolor{gray!30}$\textbf{0.026}$ & \cellcolor{gray!30}$0.95$ & \cellcolor{gray!30}$\textbf{0.969}$ & \cellcolor{gray!30}$\textbf{0.218}$ & \cellcolor{gray!30}$0.256$ & \cellcolor{gray!30}$\textbf{0.152}$ & \cellcolor{gray!30}$0.228$ \\ 
\cellcolor{lightgray!10}$2006$ & \cellcolor{lightgray!10}$\textbf{0.989}$ & \cellcolor{lightgray!10}$\textbf{0.989}$ & \cellcolor{lightgray!10}$\textbf{0.026}$ & \cellcolor{lightgray!10}$0.038$ & \cellcolor{lightgray!10}$\textbf{0.993}$ & \cellcolor{lightgray!10}$0.992$ & \cellcolor{lightgray!10}$\textbf{0.026}$ & \cellcolor{lightgray!10}$\textbf{0.026}$ & \cellcolor{lightgray!10}$0.98$ & \cellcolor{lightgray!10}$\textbf{0.985}$ & \cellcolor{lightgray!10}$0.569$ & \cellcolor{lightgray!10}$\textbf{0.397}$ & \cellcolor{lightgray!10}$\textbf{0.109}$ & \cellcolor{lightgray!10}$0.268$ \\ 
\cellcolor{gray!30}$2014$ & \cellcolor{gray!30}$0.957$ & \cellcolor{gray!30}$\textbf{0.983}$ & \cellcolor{gray!30}$0.063$ & \cellcolor{gray!30}$\textbf{0.029}$ & \cellcolor{gray!30}$0.912$ & \cellcolor{gray!30}$\textbf{0.972}$ & \cellcolor{gray!30}$\textbf{0.007}$ & \cellcolor{gray!30}$0.013$ & \cellcolor{gray!30}$0.991$ & \cellcolor{gray!30}$\textbf{0.996}$ & \cellcolor{gray!30}$\textbf{0.872}$ & \cellcolor{gray!30}$0.993$ & \cellcolor{gray!30}$\textbf{0.12}$ & \cellcolor{gray!30}$0.339$\\
\cellcolor{lightgray!10}$2021$ & \cellcolor{lightgray!10}$0.936$ & \cellcolor{lightgray!10}$\textbf{0.991}$ & \cellcolor{lightgray!10}$\textbf{0.594}$ & \cellcolor{lightgray!10}$1.0$ & \cellcolor{lightgray!10}$0.818$ & \cellcolor{lightgray!10}$\textbf{0.969}$ & \cellcolor{lightgray!10}$\textbf{0.012}$ & \cellcolor{lightgray!10}$0.026$ & \cellcolor{lightgray!10}$0.987$ & \cellcolor{lightgray!10}$\textbf{0.999}$ & \cellcolor{lightgray!10}$\textbf{0.859}$ & \cellcolor{lightgray!10}$1.0$ & \cellcolor{lightgray!10}$\textbf{0.168}$ & \cellcolor{lightgray!10}$0.314$ \\
\cellcolor{gray!30}Rural & \cellcolor{gray!30}$0.904$ & \cellcolor{gray!30}$\textbf{0.975}$ & \cellcolor{gray!30}$\textbf{0.100}$ & \cellcolor{gray!30}$0.250$ & \cellcolor{gray!30}$0.867$ & \cellcolor{gray!30}$\textbf{0.967}$ & \cellcolor{gray!30}$\textbf{0.083}$ & \cellcolor{gray!30}$0.222$ & \cellcolor{gray!30}$0.952$ & \cellcolor{gray!30}$\textbf{0.998}$ & \cellcolor{gray!30}$\textbf{0.286}$ & \cellcolor{gray!30}$1.0$ & \cellcolor{gray!30}$\textbf{0.360}$ & \cellcolor{gray!30}$0.713$ \\ 
\cellcolor{lightgray!10}Urban & \cellcolor{lightgray!10}$0.926$ & \cellcolor{lightgray!10}$\textbf{0.986}$ & \cellcolor{lightgray!10}$\textbf{0.100}$ & \cellcolor{lightgray!10}$0.263$ & \cellcolor{lightgray!10}$0.894$ & \cellcolor{lightgray!10}$\textbf{0.981}$ & \cellcolor{lightgray!10}$\textbf{0.091}$ & \cellcolor{lightgray!10}$0.238$ & \cellcolor{lightgray!10}$0.965$ & \cellcolor{lightgray!10}$\textbf{0.999}$ & \cellcolor{lightgray!10}$\textbf{0.286}$ & \cellcolor{lightgray!10}$1.0$ & \cellcolor{lightgray!10}$\textbf{0.367}$ & \cellcolor{lightgray!10}$0.722$ \\ 
\hline
\end{tabular}}
\caption{Accuracy $Acc$, Relative size $S_{rel}$ and Relative overestimation $O_{rel}$ for different Middlebury and satellite datasets. ``Global'' column consider every intervals in the dataset, while ``High confidence areas'' and ``Low confidence areas'' separate the intervals based on the confidence measure (\cref{eq:low_confidence}). Two cost functions are compared: CENSUS and MC-CNN. The best results for each dataset appear in bold font.}\label{table:results}
\end{table*}

We define different metrics in order to simultaneously evaluate intervals reliability and size. The metrics are adapted to our future objective of propagating the disparity intervals into height intervals for 3D reconstruction from satellite imagery. As such, intervals must be both reliable and small. The metrics are first assessed globally by considering every intervals for each dataset. Then, they are measured separately for high confidence and low confidence areas. The distinction between both regions is performed using a threshold on the ambiguity measure, as in \cref{eq:low_confidence}.

\subsection{Evaluation Metrics}
\begin{figure}[ht]
    \centering\includegraphics[height=4cm]{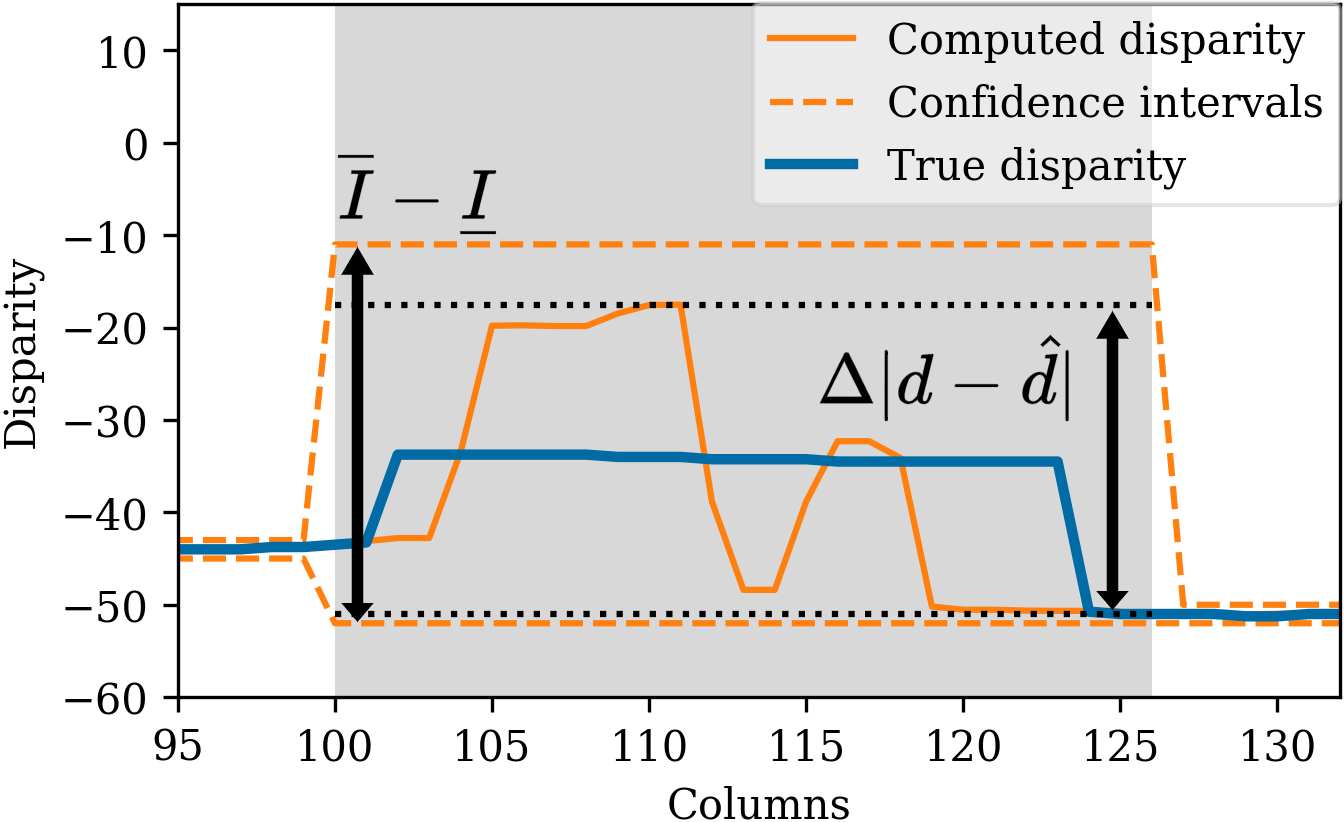}
    \caption{A low confidence area in gray, with a representation of $\Delta|d-\hat{d}|$ and $\overline{I}-\underline{I}$ from \cref{eq:relative_error}.}
    \label{fig:relative_error}
\end{figure}

The confidence intervals are evaluated following different criteria:
\begin{itemize}
    \item Their global \textit{accuracy}. An interval is considered accurate if it contains the true disparity. The accuracy is computed as:
    \begin{equation}
        Acc=\frac{\# accutate~intervals}{\# intervals}\label{eq:accuracy}
    \end{equation}
    where $\#$ refers to the number of elements of a set. 
    \item The \textit{relative size} of the intervals compared to the disparity range. The relative size is computed over a scene or a whole dataset as:
    \begin{equation}
    S_{rel}=median\left(\frac{\overline{I}-\underline{I}}{d_{max}-d_{min}}\right)\label{eq:size_of_interval}
    \end{equation}
    This criterion is important as one could achieve $100\%$ accuracy by simply setting every interval to $[d_{min}, d_{max}]$. 
    \item $S_{rel}$ is not adapted in low confidence areas as we purposely extended the intervals (see \cref{{subsec:regularization}}). Thus, we define an additional criterion only for low confidence areas, called \textit{relative overestimation}:
    \begin{equation}
        O_{rel}=median\left(1-\frac{\Delta|d-\hat{d}|}{\overline{I}-\underline{I}}\right)\label{eq:relative_error}
    \end{equation} where $\Delta|d-\hat{d}|$ is the maximal difference between the true disparity and the predicted disparity over the low confidence area. It is therefore the size of the optimal interval in the area. \Cref{fig:relative_error} allows to visualize $\Delta|d-\hat{d}|$ and $\overline{I}-\underline{I}$. $O_{rel}$ is the median of intervals overestimation over low confidence areas.
\end{itemize}

In \cref{eq:size_of_interval} and \cref{eq:relative_error}, the median is used to evaluate the sizes of the intervals instead of the mean in order to gain statistical robustness. It is noteworthy that using the mean yields very similar results. 

 In the absence of any other method for creating disparity confidence intervals, we compare the accuracy and relative size of our method with a ``naive'' approach that serves as a baseline. This approach, referred to as \textit{baseline} in \cref{tab:baseline_alpha}, consists in simply normalizing every cost curve with its maximum and minimum values, and defining the disparity interval as the minimal and maximal disparities for which the cost is greater that $90\%$. We compare this to our method using possibilities without regularization and with $\alpha$-levels taking values in $[50\%,80\%,90\%,98\%]$. We also present our method with the regularization step and an $\alpha$-level of $90\%$, referred to as ``$90$ w/ reg''. This is the same method used in \cref{table:results}. This ablation study enables us to observe the impact of different $\alpha$-levels on the accuracy of the intervals, and proves the necessity of the regularization step. 

\Cref{sec:accuracy} discusses the performance of the method with regard to the criterion from \cref{eq:accuracy}. We aim to validate at least $90\%$ accuracy on the intervals over every scene. \Cref{sec:size} evaluates criteria of \cref{eq:size_of_interval} and \cref{eq:relative_error}. In high confidence areas, the relative size $S_{rel}$ needs to be as small as possible. We consider a relative size of around $25\%$ as a satisfying objective. In low confidence areas, a relative overestimation $O_{rel}$ of about $30\%$ seems a feasible objective, while still providing enough information for a later propagation into elevation intervals. The numerical objectives are given for information purposes, as users needs may vary depending on the application.

\begin{table}[!ht]
    \centering
    \begin{tabular}{|c?{1.5pt}c|c|c|c|c|c?{1.5pt}}
        \cline{2-7}
         \multicolumn{1}{c?{1.5pt}}{} & Baseline & $50$ & $80$ & $90$ \fontsize{8}{9.6}\selectfont & $90$ \fontsize{8}{9.6}\selectfont w/ reg & $98$ \\ \hline\hline
        \cellcolor{lightgray!10}\small2003 & \cellcolor{lightgray!10}0.503 & \cellcolor{lightgray!10}\textbf{0.997} & \cellcolor{lightgray!10}0.985 & \cellcolor{lightgray!10}0.982 & \cellcolor{lightgray!10}0.973 & \cellcolor{lightgray!10}0.98\\ \hline
        \cellcolor{gray!30}\small2005 & \cellcolor{gray!30}0.592 & \cellcolor{gray!30}\textbf{0.978} & \cellcolor{gray!30}0.958 & \cellcolor{gray!30}0.950 & \cellcolor{gray!30}0.963 & \cellcolor{gray!30}0.944\\ \hline
        \cellcolor{lightgray!10}\small2006 & \cellcolor{lightgray!10}0.626 & \cellcolor{lightgray!10}0.986 & \cellcolor{lightgray!10}0.982 & \cellcolor{lightgray!10}0.979 & \cellcolor{lightgray!10}\textbf{0.989} & \cellcolor{lightgray!10}0.976\\ \hline
        \cellcolor{gray!30}\small2014 & \cellcolor{gray!30}0.039 & \cellcolor{gray!30}0.543 & \cellcolor{gray!30}0.519 & \cellcolor{gray!30}0.494 & \cellcolor{gray!30}\textbf{0.957} & \cellcolor{gray!30}0.474\\ \hline
        \cellcolor{lightgray!10}\small2021 & \cellcolor{lightgray!10}0.03 & \cellcolor{lightgray!10}0.549 & \cellcolor{lightgray!10}0.478 & \cellcolor{lightgray!10}0.442 & \cellcolor{lightgray!10}\textbf{0.936} & \cellcolor{lightgray!10}0.420\\ \hline
    \end{tabular}
    \caption{Ablation study. Evaluation of the accuracy from \cref{eq:accuracy} with different methods using the CENSUS cost function. $50,80,90,98$ refer to different value of $\alpha$. $90$ /w reg refers to an $\alpha$ value of $90\%$ and a regularization step from \cref{subsec:regularization}.}\label{tab:baseline_alpha}
\end{table}

\subsection{Reference Datasets}
We used $83$ scenes from Middlebury $2003, 2005, 2006, 2014$ and $2021$ datasets for evaluation \cite{scharstein_taxonomy_2001, scharstein_high-accuracy_2003, scharstein_learning_2007, jiang_high-resolution_2014}. We use quarter-size and third-size versions of the data for $2003, 2005$ and $2006$ datasets and full resolution for $2014$ and $2021$ datasets. We use the disparity range indicated in the calibration files. Each year contains respectively $2$, $6$, $21$, $23$ and $24$ pairs of images with different shapes, and the size of the disparity intervals ranges between $60$ and $1110$. We also use $120$ $1845\times1845$ pairs of epipolar images generated using \cite{cournet_ground_2020} from satellite images of the region of Montpellier, France, with a resolution of $50$cm/pixel. The disparity range for those images is between $20$ and $50$ pixels, depending on the scene. During the evaluation, the dataset is split into two categories: \textit{urban}, for images containing mostly buildings, and \textit{rural}, for image mostly composed of forests and fields. The ground truth disparity was retrieved using LiDAR data.

\subsection{Accuracy Results}\label{sec:accuracy}
\begin{figure}[ht]
\centering\begin{subfigure}{0.5\linewidth}
  \includegraphics[width=1\linewidth]{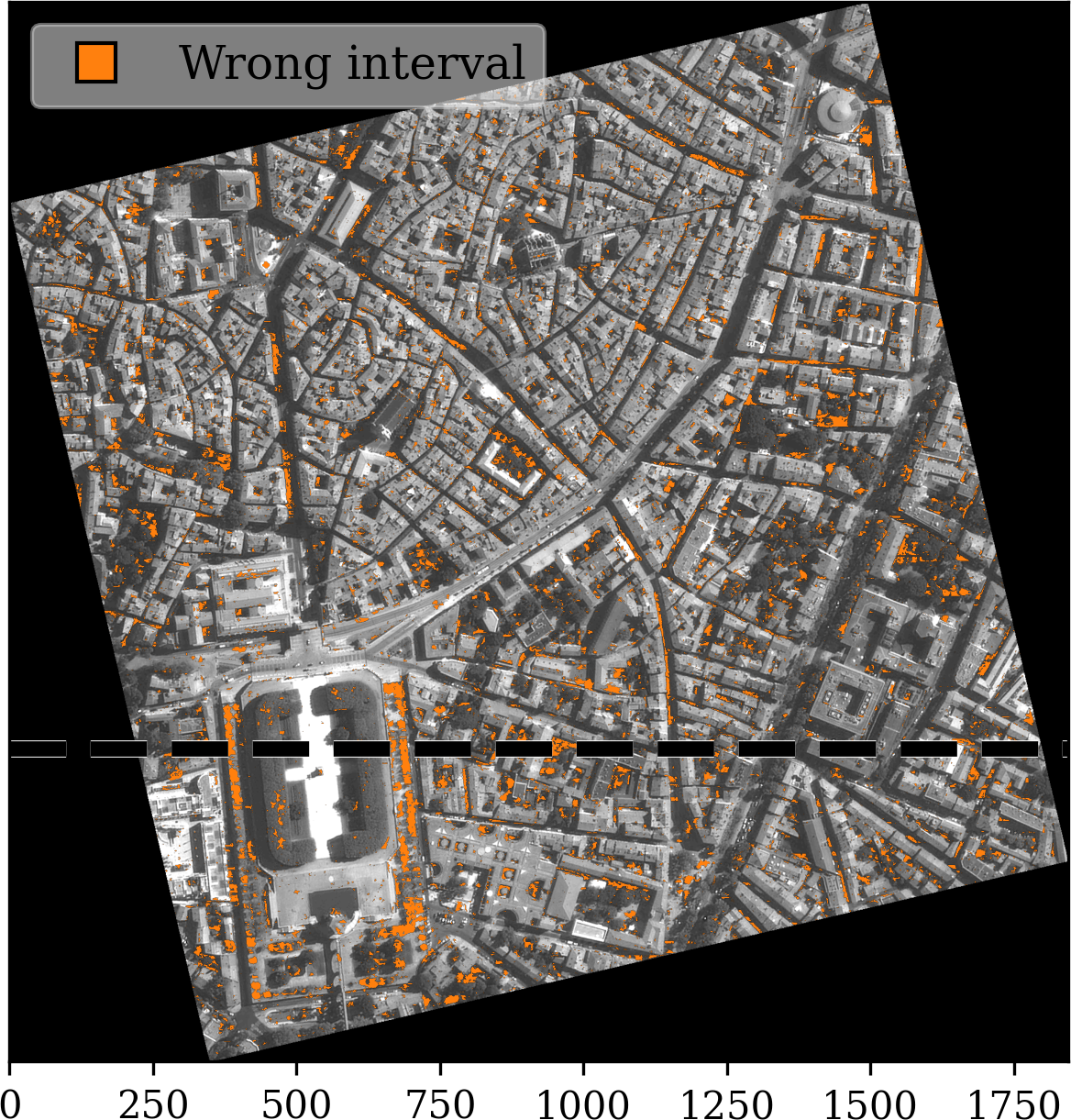}
  \caption{Left epipolar image}
    \label{fig:fig_7_a}
\end{subfigure}
\begin{subfigure}{1\linewidth}
    \centering
  \includegraphics[width=1\linewidth]{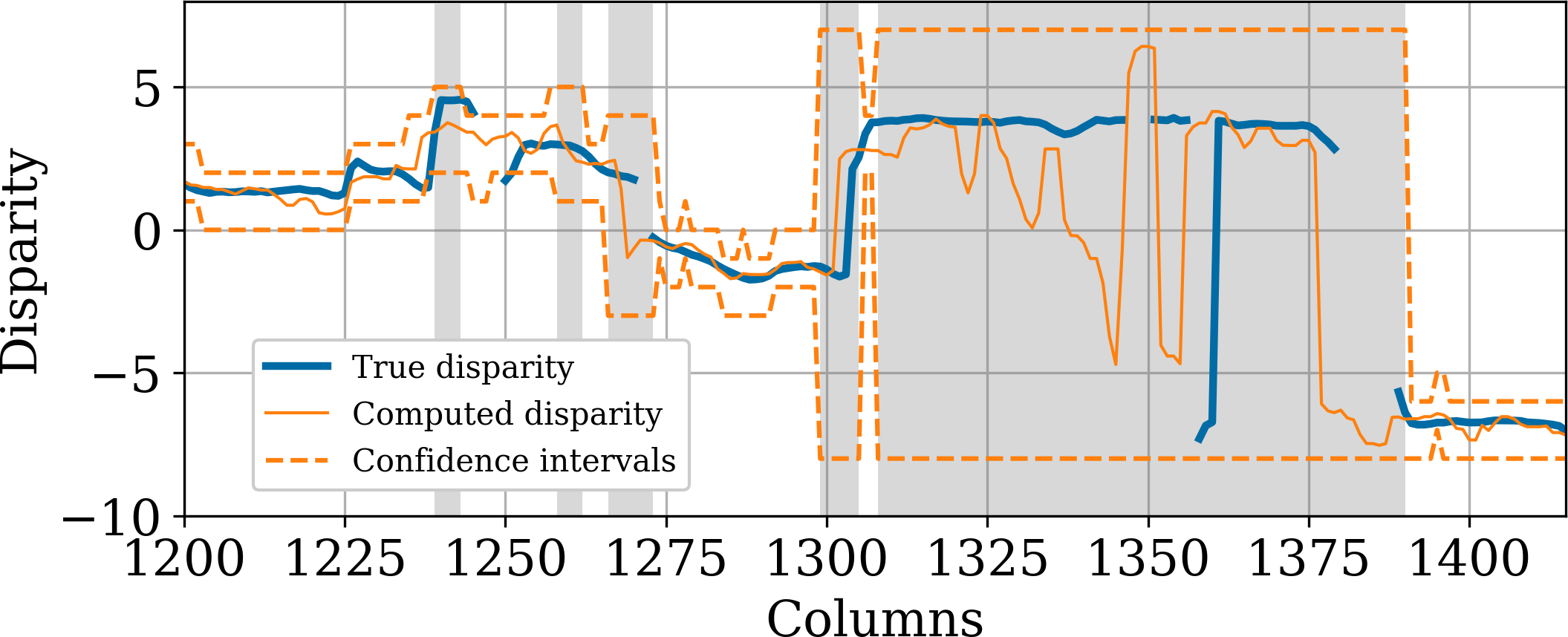}
    \caption{Confidence Intervals}
  \label{fig:fig_7_b}
\end{subfigure}
\begin{subfigure}{1\linewidth}
    \centering
    \includegraphics[width=1\linewidth]{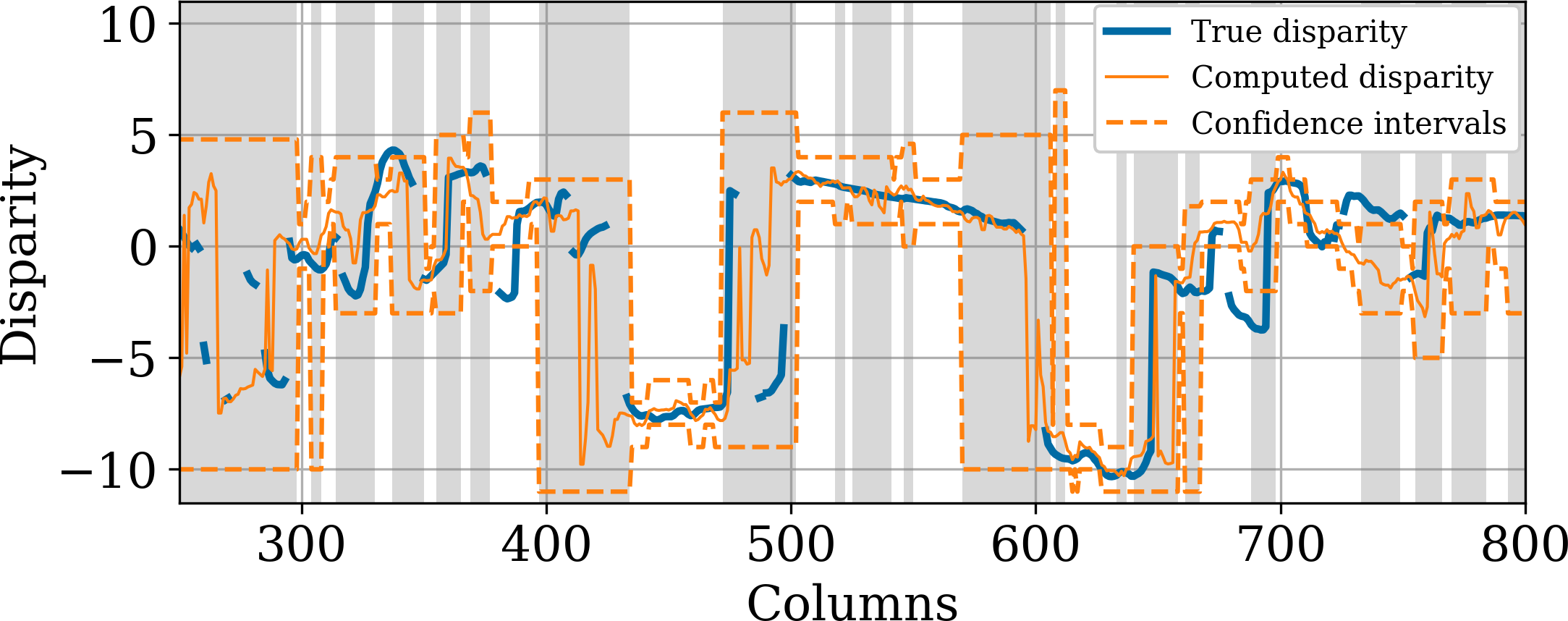}
    \caption{Confidence Intervals}
  \label{fig:fig_7_c}
\end{subfigure}
\caption{\cref{fig:fig_7_a} is the left image of the city of Montpellier where colored pixels indicate wrong interval location. \cref{fig:fig_7_b,fig:fig_7_c} present detailed confidence intervals, computed disparity, and true disparity along the black dashed line from figure \cref{fig:fig_7_a}. Areas with low confidence are indicated in gray.}\label{fig:fig_7}
\end{figure}

First, intervals accuracy can be analyzed on the specific example of \cref{fig:fig_7_a}. Inaccurate intervals are colored in the left image. \Cref{fig:fig_7_b,fig:fig_7_c} present confidence interval values along a row, as well as the disparity estimation and the true disparity. Low confidence areas are indicated by the gray sections. \cref{fig:fig_7_b,fig:fig_7_c} contain both high confidence areas with small intervals, and low confidence areas with important disparity variations. We observe a low confidence area between columns $1300$ and $1390$ where the computed disparity is far from the true disparity, but the confidence intervals remain correct. 
 
Evaluating the accuracy statistics on each dataset yields strong results. Scores per year for intervals computed using the CENSUS and MC-CNN cost functions are presented in \cref{table:results}. CENSUS-based intervals have an accuracy always superior to $90\%$ over each dataset. They verify the $90\%$ accuracy objective on $80$ of the $83$ scenes from Middlebury, and on all $120$ satellite images. MC-CNN-based intervals have an accuracy superior to $95\%$ on every datasets. They validate the $90\%$ accuracy objective on all $83$ scenes from Middlebury and $120$ satellite images. Those strong performances come nonetheless with large interval size, as detailed in section \cref{sec:size}.

An ablation study is carried out in \cref{tab:baseline_alpha}, highlighting the importance of the regularization process. Without the regularization, the accuracy drops for datasets with many low confidence areas. The value of $\alpha$ has a small impact when compared to the regularization step. We also observe that the naive approach of the baseline produces very inaccurate intervals in comparison to our method.

\subsection{Discussions on the Size of the Intervals}\label{sec:size}
\begin{figure}[ht]
\centering\begin{subfigure}{0.5\linewidth}
  \includegraphics[width=1\linewidth]{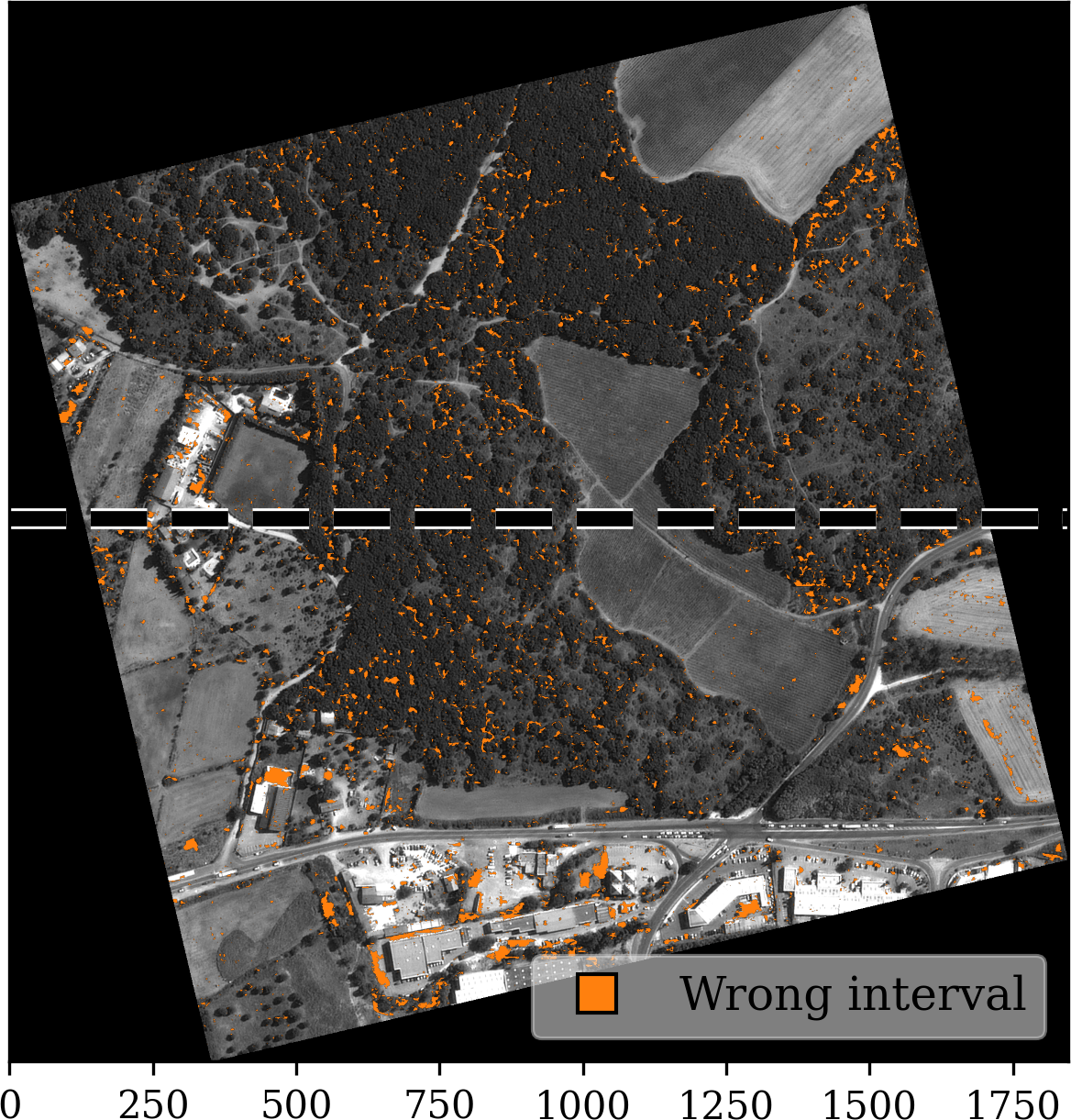}
  \caption{Left epipolar image}
    \label{fig:fig_8_a}
\end{subfigure}
\begin{subfigure}{1\linewidth}
    \centering
  \includegraphics[width=1\linewidth]{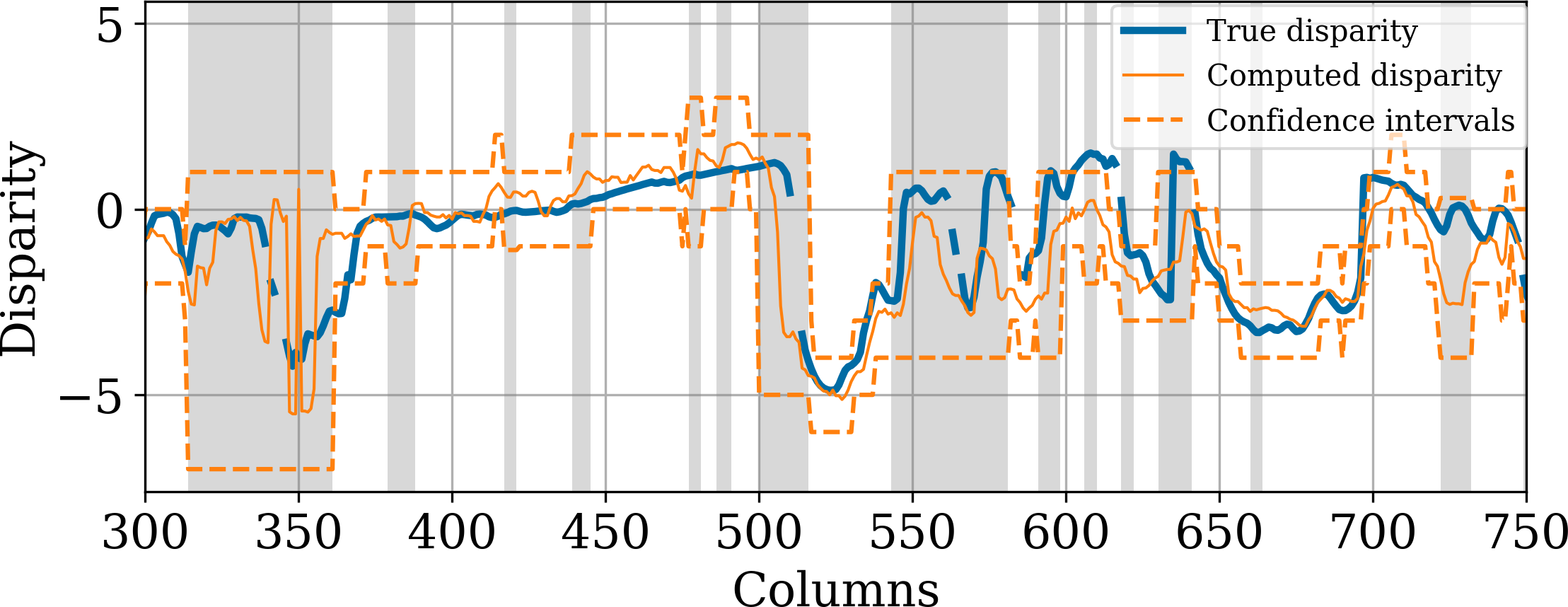}
    \caption{Confidence Intervals}
  \label{fig:fig_8_b}
\end{subfigure}
\caption{\cref{fig:fig_8_a} is the left image of a rural area near Montpellier. Colored pixels indicate wrong interval locations. \Cref{fig:fig_8_b} details confidence intervals, computed disparity, and true disparity along the black dashed line from figure \cref{fig:fig_8_a}. Areas with low confidence are indicated in gray.}
   \label{fig:fig_8}
\end{figure}

Although the intervals are very accurate, confidence intervals with unnecessarily large sizes need to be avoided. In \cref{fig:fig_7_b,fig:fig_7_c}, confidence intervals have a small relative size in high confidence areas, and a larger size in low confidence areas. In those areas, the intervals are properly adjusted to contain the predicted disparity and the true disparity without overestimating the error. Similar observations can be made in \cref{fig:fig_8_b}.

Detailed statistics of intervals relative size and relative overestimation are presented in \cref{table:results} alongside accuracy results. For MC-CNN-based intervals, the $5\% $ relative size criterion is validated for each dataset in high confidence areas, with a relative size on all datasets of only $1.5\% $. The global relative size is below $4\% $ for years $2001, 2003, 2005, 2006$ and $2014$. $S_{rel}$ is maximal for the $2021$ dataset due to the high proportion of low confidence intervals on this dataset. Intervals in low confidence areas are only overestimated by around $30\% $, meaning that large intervals are unavoidable on this dataset. This can also be explained by the complexity and high resolution of $2021$ (and $2014$) scenes, which leads to larger confidence intervals in general. It results in poorer performances of the disparity prediction as a majority of pixels have low confidence. Intervals computed on satellite images have a relative size around $25\%$, which is relatively low as most scenes have a disparity range of around $20$ pixels. They however tend to overestimate the intervals in low confidence areas: around $30\%$ when using the CENSUS cost function, and $70\%$ when using the MC-CNN cost function.

Intervals computed using the CENSUS cost function validate the $25\% $ relative size objective in high confidence areas. In low confidence areas, their relative overestimation is around $13\% $, meaning that they are close to the ideal intervals. They outperform MC-CNN based intervals on all datasets when comparing their relative size in high confidence, and on all datasets regarding the relative overestimation criterion. This is probably due to the SGM regularization, which uses different weight for the CENSUS and MC-CNN cost functions. CENSUS based weights seem to produce curves with a more pronounced/narrow peak near the minima, resulting in smaller intervals, and thus better relative sizes and over-estimation metrics.
\section{Conclusion and Perspectives}
To the best of our knowledge, we present the first method for creating confidence intervals on the disparity in stereo matching problems. Our method is designed to work with any stereo algorithm computing a 3D cost volume. Matching cost functions are transformed into possibility distributions and then interpreted as an expert's opinion. We rely on the advanced theoretical background of possibility distributions to compute robust uncertainty estimations. Confidence intervals are deduced from the $\alpha$-cuts of possibility distributions, and regularized in low confidence areas. Post-processing steps handling is also taken into account to maintain consistency between the predicted disparity map and the confidence intervals.  As we have not found existing accurate methods for comparison, we assess the intervals based on accuracy, relative size, and overestimation. Criteria are evaluated on the Middlebury datasets. $90\% $ accuracy objective is achieved while maintaining a small relative size in high confidence area and without overestimating the intervals size in low confidence areas. The accuracy of the confidence intervals does not depend on the performance of the disparity estimation. All of our contributions are available on our GitHub repository. 
This work aims to motivate further research in detecting, locating and quantifying the magnitude of the error in disparity maps. We demonstrate in this paper that possibility distributions can model and process epistemic uncertainty in a understandable and explainable way.

Future work will include propagating the disparity confidence intervals into elevation confidence intervals for 3D reconstruction. Those intervals can then be provided as a complementary product alongside digital surface models, often used in many Earth Observation applications.
{
    \small
    \bibliographystyle{ieeenat_fullname}
    \bibliography{main}
}

\end{document}